# A Comparative Study of YOLOv8 to YOLOv11 Performance in Underwater Vision Tasks

Gordon Hung[1], Ivan Felipe Rodriguez[2]

[1]Department of Computer Science, Hsinchu County American School, Taiwan
[2]Department of Cognitive Sciences, Brown University, Rhode Island, USA

Corresponding author: Gordon Hung (gordonh741@gmail.com).

**ABSTRACT** Autonomous underwater vehicles (AUVs) increasingly rely on on-board computer-vision systems for tasks such as habitat mapping, ecological monitoring, and infrastructure inspection. However, underwater imagery is hindered by light attenuation, turbidity, and severe class imbalance, while the computational resources available on AUVs are limited. One-stage detectors from the YOLO family are attractive because they fuse localization and classification in a single, low-latency network; however, their terrestrial benchmarks (COCO, PASCAL-VOC, Open Images) leave open the question of how successive YOLO releases perform in the marine domain. We curate two openly available datasets that span contrasting operating conditions: a Coral Disease set (4,480 images, 18 classes) and a Fish Species set (7,500 images, 20 classes). For each dataset, we create four training regimes (25 %, 50 %, 75 %, 100 % of the images) while keeping balanced validation and test partitions fixed. We train YOLOv8-s, YOLOv9-s, YOLOv10-s, and YOLOv11-s with identical hyperparameters (100 epochs, 640 px input, batch = 16, T4 GPU) and evaluate precision, recall, mAP50, mAP50-95, per-image inference time, and frames-per-second (FPS). Post-hoc Grad-CAM visualizations probe feature utilization and localization faithfulness. Across both datasets, accuracy saturates after YOLOv9, suggesting architectural innovations primarily target efficiency rather than accuracy. Inference speed, however, improves markedly. Our results (i) provide the first controlled comparison of recent YOLO variants on underwater imagery, (ii) show that lightweight YOLOv10 offers the best speed-accuracy trade-off for embedded AUV deployment, and (iii) deliver an open, reproducible benchmark and codebase to accelerate future marine-vision research.

## I. INTRODUCTION

In recent years, the advancements in autonomous underwater vehicles have led to greater demands for powerful computer vision models capable of producing real-time results [1]. Contrary to the traditional land-based tasks like facial recognition and traffic analysis, underwater object detection poses a new set of challenges [2, 3]. Images taken in the water often suffer from light distortions and turbidity, making objects appear blurry and thus harder to identify [4]. Furthermore, many underwater objects, like marine organisms, are small and often densely packed together, making it difficult for models to detect them [5]. Lastly, underwater tasks often lack large, high-quality datasets for model training, significantly hindering the model's accuracy [6].

Therefore, machine learning models used in AUVs for marine tasks need to be both accurate and highly efficient. They need to be flexible with different amounts of training data and quick at producing real-time inference [7]. Due to this requirement, many traditional two-stage detectors like the Faster R-CNN, though accurate, are too computationally heavy to deploy in AUVs [8]. Therefore, the YOLO (You Only Look Once) model has gained widespread popularity because of its combination of detection and classification modules, significantly reducing inference speed and improving accuracy. Its unified architecture enables real-time performance, making it ideal for time-sensitive applications such as autonomous navigation and live monitoring [9].

While the YOLO model is a significant breakthrough in computer vision, its existing benchmarks are all based on terrestrial imaging [10]. Common benchmark datasets include COCO, PASCAL VOC, and Open Images, none of which include any marine objects or underwater imagery. Due to the significant difference in terrestrial and marine computer vision tasks, it is unsuitable to use a single benchmark for both cases.

This leaves the relative performances of each of the YOLO versions unexplored in the field of marine science.

## II. Related Work

Underwater object detection has become increasingly vital due to its wide range of applications in marine ecology, fisheries, and conservation efforts [11 - 15]. Traditional sensing techniques like sonar have been utilized for decades for object detection, but their limited resolution and sensitivity to noise limit their effectiveness significantly. Lu et al. explain that optical imaging sensors are not replaceable for near-surface underwater operations, but also emphasize problems like light attenuation, scattering, and geometrical distortion [16]. Therefore, their study suggests experimentation with deeper computer vision models to improve performance. Some articles employ YOLO for specific underwater detection use cases. Kuswantori et al. introduce a YOLOv4 pipeline optimized for industrial fish detection, achieving 98.15% accuracy on high-speed video streams of swimming fish [17]. Their study emphasizes the potential of implementing automated sorting for fish and other industries. Similarly, Muksit et al. introduce YOLO-Fish, with two lean variations adapted for complex habitats and small-object detection, outperforming YOLOv3 on the DeepFish and OzFish datasets while achieving similar efficacies as YOLOv4 [18].

These papers validate YOLO's effectiveness in handling underwater detections. However, the current literature focuses more on task-level analyses rather than version-to-version comparisons. And for papers that do present analyses across YOLO models, their focus is primarily centered around terrestrial objects. Maity et al. compare YOLO versions v5-8 to detect terrestrial vehicles and conclude that YOLOv7 was superior on IRUVD (mAP 0.96), whereas YOLOv8 was superior on JUVDsi v1, highlighting YOLO's reliance on the dataset to generate predictions [19]. Similarly, Kim et al. compare YOLO, SSD, and Faster R-CNN to detect vehicle types, where YOLOv4 outperforms the others with 93% accuracy [20]. Cheng illustrates a larger comparison of multi-object detectors based on CNN, where YOLO is represented as being efficient [21]. His study emphasizes the architectural strength of YOLO for quickly identifying densely located objects. Terven et al. and Hussain survey YOLO's development from v1 through to v8, charting architectural developments, training techniques, and performance trade-offs in applications as varied as robotics to autonomous vehicles and defect inspection in manufacturing [22, 23]. Tyagi et al. focus on wildlife detection, outlining YOLO's application to monitoring a variety of species [24]. However, these studies do not cover state-of-the-art YOLO versions from v9 to v11, nor do they cover their applications in underwater settings.

Together, these papers indicate a major gap: while YOLO has been successfully transferred to underwater detection in single-use applications (e.g., fish sorting, reef monitoring), no serious benchmarking of sequential YOLO releases on oceanic datasets has occurred.

This work closes this gap by (i) evaluating YOLOv8–YOLOv11 on two public underwater datasets in coral disease classification and fish species detection, (ii) faithfully comparing accuracy and performance across versions under various data conditions, and (iii) employing Grad-CAM visualizations to study feature utilization and improve model selection interpretability.

## III. Methodology

### A. Datasets

This paper utilizes two marine-based datasets to analyze the performance of YOLOv8-11 in underwater settings. Since both datasets are open-access databases, each sample was verified manually to ensure robust training. The first dataset contains 4,480 images of coral diseases with 18 classes [25]. The second dataset is almost twice as large to simulate a different set of data conditions; it contains 7,500 images of fish types with 20 distinct classes [26]. Both datasets have a roughly equal number of samples per class, helping to prevent model bias.

Each dataset is balanced and preprocessed before model training. The validation and test sets of both datasets are rearranged so that each class contains exactly 20 samples, preventing potential overfitting. The training sets, on the other hand, consist of all the other images. Furthermore, this study seeks to comprehensively evaluate each YOLO model under varying dataset conditions. Therefore, we made four splits on the training set to understand how the abundance or scarcity of the dataset may affect each version of the model differently. The split is 100 percent, 75 percent, 50 percent, and 25 percent, with no modifications to both the validation and testing sets. Figures 1 and 2 show the total number of samples in each class for the two datasets.

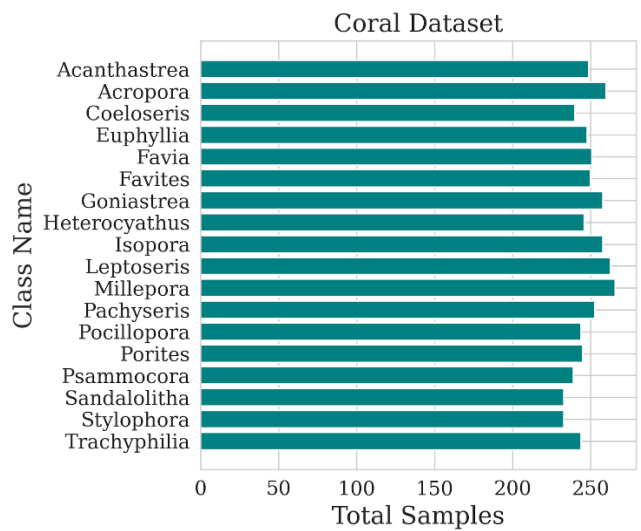

FIGURE 1. Total number of samples per class in the Coral Dataset.

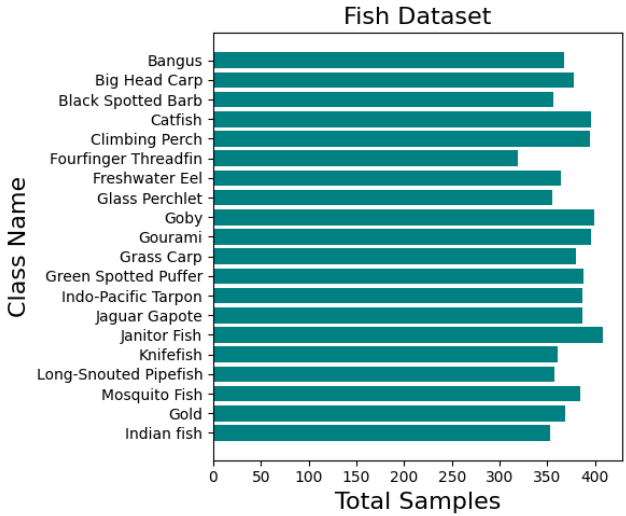

FIGURE 2. Total number of samples per class in the Fish Dataset.

## B. Metrics

To comprehensively evaluate each model, we employed a variety of evaluation metrics to understand the classification confidence, detection accuracy, and speed of inference. Table 1 includes the metrics we considered.

TABLE I
METRICS AND DEFINITIONS

| Metric | Definition |
|---|---|
| Precision | Proportion of correctly detected objects among all detections. |
| Recall | Proportion of correctly detected objects among all ground truth objects. |
| mAP@0.5 | Mean Average Precision at IoU threshold 0.5. |
| mAP50-95 | Mean Average Precision averaged over IoU thresholds from 0.5 to 0.95. |
| Time per Image | Average inference time required to process one image. |
| FPS | Number of images processed per second. |

## C. YOLO Family

Traditional object detection techniques often rely on a two-stage machine learning pipeline where one model is used for object identification and another is employed for object classification. This method has been proven to be both computationally heavy and incredibly slow. Therefore, Joseph Redmon et al introduced the YOLO model that revolutionized computer vision. YOLO combines the aforementioned two-stage pipeline into a single model, significantly improving the efficiency for detection tasks.

The YOLO model only inspects an image once and divides it into small grid cells. Each cell is responsible for calculating the likelihood that it contains an object and the confidence score for that prediction. This core concept allows the model to be highly efficient and robust.

However, though ground-breaking, the YOLO model has several limitations. First, because it uses each grid cell to generate predictions, it is difficult for the model to recognize two objects in the same cell. Furthermore, the accuracy of the YOLO model is often inferior to that of the traditional two-stage pipeline. Regardless, it still remains one of the most powerful and popular computer vision models.

Since its inception, YOLO has undergone various upgrades, as seen in Figure 3. This paper focuses on the four state-of-the-art YOLO models, namely YOLOv8, YOLOv9, YOLOv10, and YOLOv11.

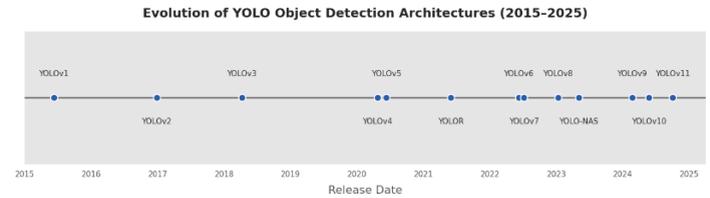

FIGURE 3. YOLO evolution timeline.

## D. YOLOv8

YOLOv8 utilizes a very different architecture compared to its predecessors [27]. Traditional YOLO models relied heavily on anchor boxes, which are predefined bounding boxes used to predict the locations and sizes of various objects. YOLOv8, on the other hand, uses an anchor-free head to improve the accuracy of detecting small objects. Furthermore, rather than using a unified head, YOLOv8 separates the regression and classification tasks. The regression task in computer vision is to recognize where the object is, whereas the classification task is simply to analyze what the object is. This decoupling allows the model to focus on each task independently, therefore improving both the classification and detection accuracy. Additionally, YOLOv8 employs Reparameterized Convolution blocks during training and simpler Convolution blocks for inference. This allows the model to accurately grasp the image patterns while reducing computational burden.

## E. YOLOv9

YOLOv9 introduces a novel hybrid design that allows for greater flexibility [28]. It uses a Unified Detection Head (UDH) to combine the strengths of anchor-based and anchor-free designs into a single hybrid detection head. This improves the accuracy by allowing the model to dynamically decide which strategy works best based on the object of interest. Furthermore, instead of the standard YOLO backbones, YOLOv9 uses the Generalized Efficient Layer Aggregation Network (GELAN) to ease the computational burden while keeping the accuracy. It allows the model to understand the context of the image more deeply without increasing the inference time. Furthermore, to adjust previous YOLO models' inability to identify objects in close proximity,

YOLOv9 introduces the Dynamic Receptive Field Selection (DRS) Block to dynamically adjust the receptive field size for each feature map layer. YOLOv9 also utilizes a refined feature pyramid structure to balance spatial and semantic information. This allows the model to detect both small and large objects effectively, a detrimental limitation of previous YOLO models.

*F. YOLOv10*

YOLOv10 is a lightweight model optimized for deployment on edge devices [29]. This design follows the increasing trend in the utilization of machine learning algorithms in autonomous vehicles. YOLOv10 uses Neural Architecture Search (NAS) to build the backbone and neck architectures. NAS is an optimization technique that finds the optimal design of the neural network through a search algorithm, thus increasing the accuracy and speed of the model. YOLOv10 also uses an improved C3 module for the neck for better feature aggregation. This allows the gradient to flow more smoothly during training, allowing the model to learn the features more effectively. Similar to YOLOv8, YOLOv10 supports a variety of computer vision tasks, including detection, segmentation, pose estimation, and tracking.

*G. YOLOv11*

YOLOv11 is the state-of-the-art architecture of the YOLO family [30]. Contrary to the traditional CNN-only architecture, YOLOv11 utilizes CNN-based backbones with attention mechanisms to capture both the local details and the global dependencies. YOLOv11 also uses a dynamic scaling mechanism that allows it to find the optimal width and depth depending on the training data. This optimizes the model for both accuracy and efficiency. Furthermore, YOLOv11 uses a novel neck design that employs lightweight transformers like EfficientFormer for quick context understanding. This allows the model to run inference at an extremely fast rate. Instead of the traditional loss function, YOLOv11 uses the distribution focal loss (DFL) for bounding box detection and task-aligned focal loss for classification. This allows the model to achieve better localization with more precise tasks.

*H. Comparison*

Table 2 presents a detailed comparison of the key architectural innovations of each version of the YOLO model. Each version is optimized to be more computationally efficient and accurate compared to the previous model.

TABLE II
KEY YOLO INNOVATIONS

| Feature | YOLOv8 | YOLOv9 | YOLOv10 | YOLOv11 |
|---|---|---|---|---|
| Backbone | Custom CNN | GELAN | NAS-searched CNN | CNN + Attention |
| Neck | Reparam Conv Blocks | Refined Feature Pyramid | Improved C3 Module | Lightweight Transformers |
| Head | Decoupled Head | United Detection Head | Decoupled Head | Task-aligned Head |
| Anchor | Anchor-free | Hybrid | Anchor-free | Anchor-free |
| Special Modules | Decoupled Structure, Reparam Conv | Dynamic Receptive Field Selection | NAS, Improved C3 | Dynamic Scaling, DFL, Attention |
| Loss Function | CIoU Loss | Improved variant | Standard + auxiliary tasks | Distribution Focal Loss, Task-aligned Focal Loss |
| Optimization Focus | Accuracy on small objects | Proximity object detection + flexibility | Lightweight edge development | Speed + contextual learning |

Each YOLO version has several model sizes. This research employs YOLO small due to its widespread popularity and efficiency. Table 3 illustrates the specific architectural differences between each model.

TABLE III
KEY MODEL DIFFERENCES

| Model | Layers | Parameters | GFLOPs |
|---|---|---|---|
| YOLO8s | 129 | 11166560 | 28.8 |
| YOLO9s | 544 | 7318368 | 27.6 |
| YOLO10s | 234 | 8128272 | 25.1 |
| YOLO11s | 181 | 9458752 | 21.7 |

*I. Training*

We trained each YOLO version from scratch with different portions of the training data to fully understand their respective performances and improvements. For example, we trained the YOLOv8 model with 25 percent, 50 percent, 75 percent, and 100 percent of the entire dataset; the same exact framework is applied to all the versions. This data partitioning framework is only applied to the training set; both the validation and test sets remain the same. This allows us to analyze how different amounts of training data affect different YOLO versions. Furthermore, by using two datasets of different scales, we can also analyze how reliant each version is on the data size. Each training process follows the standard training convention. The training parameters are shown in Table 4.

TABLE IV
CONFIGURATIONS

| Training Parameters | Values |
|---|---|
| Epochs | 100 |
| Image Size | 640 |
| Batch Size | 16 |
| Device | T4 GPU |
| Workers | 4 |
| Optimizer | auto |

*J. Grad-CAM*

To understand how each model generates predictions, we employed Grad-CAM visualizations to highlight key features.

This technique allows greater transparency and interpretability of state-of-the-art YOLO models. While explainability is fairly simple in traditional classification models like standard Convolutional Neural Networks (CNNs), it is highly difficult for a two-function model that allows for real-time detection and classification. Therefore, we built upon Borah et al.'s research on developing an open-source YOLOv8 Explainer that uses the post hoc explainability technique to generate heatmaps to locate key features [31]. We manually adjusted the program configurations to use different layers for Grad-CAM, adjusting to the varying architectures of each YOLO model. We applied the Grad-CAM technique to every model. Figure 4 explains the workflow of Grad-CAM.

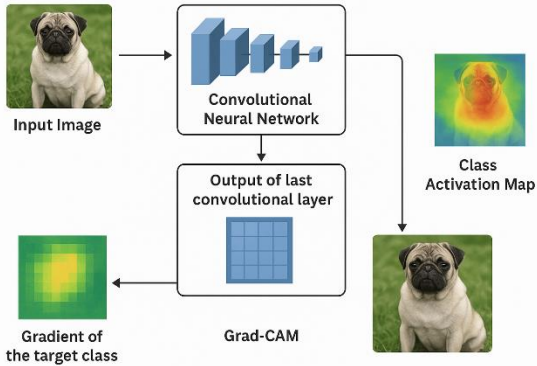

FIGURE 4. Grad-CAM overview

The process begins with an input image. The CNN processes the image with convolutional layers, producing activation maps. Then, these maps are passed through fully connected layers and processed to get a score vector for each class. The higher the score, the more likely the model thinks the image belongs to that specific class. The gradient of the score for the target class is then calculated with the activation map to understand how each pixel contributed to the prediction. These gradients are then averaged to get a weight for each feature map, thus indicating how important they are for the class. The result is a heatmap shown in red, yellow, and blue. The closer the color is to red, the more important the region is to the final prediction.

## IV. Results and Analysis

Upon careful experimentation and evaluation, our results show several limitations of the state-of-the-art YOLO models. First, we analyzed the performances of the models in the Coral Dataset. Table 5 presents the performances of each of the models trained on different sets of the dataset, with the best-performing models highlighted.

TABLE V
RESULTS OF YOLO MODELS ON THE CORAL DATASET

| Model | Precision | Recall | mAP50 | mAP50-95 |
|---|---|---|---|---|
| YOLOv11_100p | 0.92931 | 0.87382 | 0.92378 | 0.6948 |
| YOLOv10_100p | 0.94925 | 0.85847 | 0.93286 | 0.70798 |
| YOLOv9_100p | 0.93346 | 0.87573 | 0.92856 | 0.70527 |
| YOLOv8_100p | 0.92903 | 0.88576 | 0.91121 | 0.67444 |
| YOLOv11_75p | 0.91599 | 0.86539 | 0.91411 | 0.66966 |
| YOLOv10_75p | 0.92364 | 0.88052 | 0.91634 | 0.67767 |
| YOLOv9_75p | 0.91865 | 0.84993 | 0.90869 | 0.6795 |
| YOLOv8_75p | 0.89179 | 0.87681 | 0.89356 | 0.65324 |
| YOLOv11_50p | 0.8742 | 0.804 | 0.85562 | 0.62471 |
| YOLOv10_50p | 0.86496 | 0.82117 | 0.86994 | 0.62367 |
| YOLOv9_50p | 0.87074 | 0.82888 | 0.87504 | 0.65293 |
| YOLOv8_50p | 0.90993 | 0.83481 | 0.87345 | 0.62647 |
| YOLOv11_25p | 0.79499 | 0.68211 | 0.7371 | 0.50946 |
| YOLOv10_25p | 0.84700 | 0.66591 | 0.75767 | 0.53558 |
| YOLOv9_25p | 0.82424 | 0.67693 | 0.75828 | 0.53128 |
| YOLOv8_25p | 0.77796 | 0.6765 | 0.73193 | 0.51173 |

As shown in Table 5, the models trained on a greater portion of the training set achieved better accuracy. This simply demonstrates that the models were not suffering from overfitting or noise in the dataset. However, the data also shows no significant difference between different versions of the YOLO model trained on the same portion of the training set. For example, for models trained on 100% of the training set, both YOLOv11 and YOLOv9 achieved a precision of around 0.93, a recall of around 0.87, an mAP50 of around 0.92, and an mAP50-95 of around 0.70. This same trend can be observed in all the training portions, with models trained on 25% of the training set demonstrating the greatest variability. These relative similarities in accuracy suggest that, despite architectural innovations, the YOLO family has not improved drastically for marine settings. Figure 5 compares the accuracies across all models across different training sets.

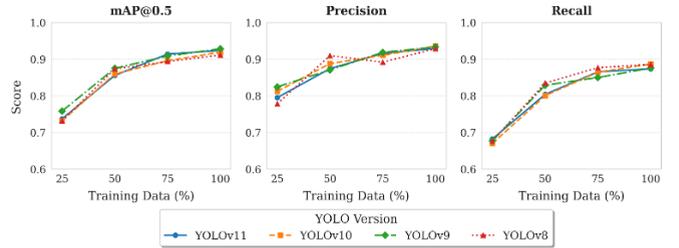

FIGURE 5. Performances of YOLOv8-11 on the Coral Dataset.

We can see from Figure 5 that there is little difference between the results of the different YOLO versions. Furthermore, we evaluated the inference speed of each model version to analyze their efficiency. Since the inference speed is independent of the training size or time, the model trained on the entire training set was utilized. The results are shown in Table 6.

TABLE VI
INFERENCE SPEED OF EACH YOLO VERSION

| Model | Time per Image (s) | FPS |
|---|---|---|
| YOLO8s | 0.0468 | 21.39 |
| YOLO9s | 0.0265 | 37.8 |
| YOLO10s | 0.0228 | 43.92 |
| YOLO11s | 0.0231 | 43.38 |

Despite the relative similarities between the accuracies of the YOLO models, Table 6 shows that the inference time has been reduced for the latest versions. However, YOLOv10 achieved a greater FPS score compared to YOLOv11, which is contradictory, as YOLOv11 is optimized for speed. A similar condition can be observed for the fish dataset. The performances of the models can be seen in Table 7.

TABLE VII
RESULTS OF YOLO MODELS ON THE FISH DATASET

| Model | Precision | Recall | mAP50 | mAP50-95 |
|---|---|---|---|---|
| YOLOv11_100p | 0.95128 | 0.93562 | 0.96869 | 0.79634 |
| YOLOv10_100p | 0.92756 | 0.96171 | 0.96901 | 0.79641 |
| YOLOv9_100p | 0.95971 | 0.92497 | 0.96774 | 0.80244 |
| YOLOv8_100p | 0.9587 | 0.92796 | 0.96958 | 0.7831 |
| YOLOv11_75p | 0.93285 | 0.93391 | 0.95825 | 0.79119 |
| YOLOv10_75p | 0.93718 | 0.91673 | 0.96142 | 0.78767 |
| YOLOv9_75p | 0.92442 | 0.93441 | 0.96439 | 0.79946 |
| YOLOv8_75p | 0.93499 | 0.93059 | 0.96196 | 0.77989 |
| YOLOv11_50p | 0.94989 | 0.90376 | 0.96489 | 0.77907 |
| YOLOv10_50p | 0.91556 | 0.90802 | 0.95368 | 0.77253 |
| YOLOv9_50p | 0.92977 | 0.87077 | 0.93858 | 0.76939 |
| YOLOv8_50p | 0.94634 | 0.86185 | 0.93806 | 0.75245 |
| YOLOv11_25p | 0.88799 | 0.8486 | 0.90963 | 0.7145 |
| YOLOv10_25p | 0.93178 | 0.84224 | 0.92118 | 0.72638 |
| YOLOv9_25p | 0.86358 | 0.86749 | 0.91935 | 0.73109 |
| YOLOv8_25p | 0.9039 | 0.81999 | 0.89696 | 0.69833 |

We can see from Table 7 that despite architectural innovations, the accuracy of the models does not seem to improve with newer versions of the model. In fact, for the majority of the sets, YOLOv8 and YOLOv9 achieved better results compared to YOLOv10 and YOLOv11. Figure 6 compares the accuracies of the models across different training sets.

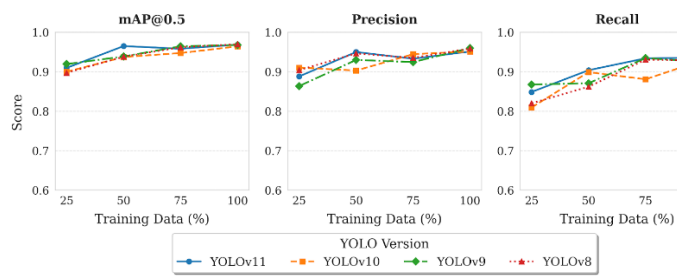

FIGURE 6. Performances of YOLOv8-11 on the Fish Dataset.

As we can see from Figure 6, the models generally follow the same trend with minimal differences. Table 8 offers comparisons of the inference time of each YOLO version.

TABLE VIII
INFERENCE SPEED OF EACH YOLO VERSION

| Model | Time per Image (s) | FPS |
|---|---|---|
| YOLO8s | 0.0620 | 16.14 |
| YOLO9s | 0.0273 | 36.6 |
| YOLO10s | 0.0230 | 43.51 |
| YOLO11s | 0.0241 | 41.41 |

We can see that YOLOv8 achieved the slowest inference time due to its heavy architectural design, whereas YOLOv10 achieved the optimal performance. However, we can see that YOLOv10 again outperformed YOLOv11 in inference speed. This result showcases that YOLOv11's architectural innovations may not be ideal in marine object detection tasks.

*Grad-CAM*

Upon analysis of the statistical results of the models, this section uses explainability techniques to further analyze each model.

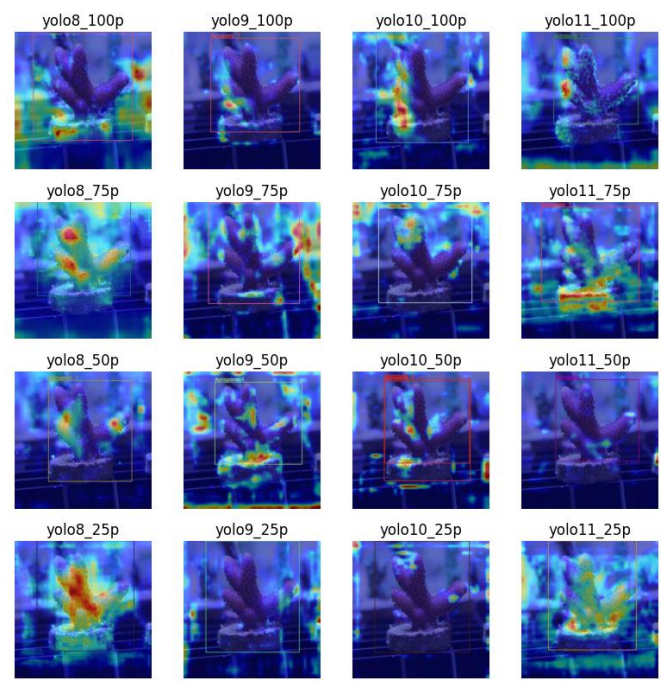

FIGURE 7. Grad-CAM from coral images.

We can see from Figure 7 that, despite the multiple advancements of the YOLO model, explainability analysis reveals that models still leverage spurious features that are not directly linked to the object to make predictions. For example, we can see that as the model is trained on more data samples, the performance does not improve significantly. For example, when YOLOv8 is trained on 25 percent of the training set, it seems to achieve better results than when it is trained on 100 percent of the dataset. These inconsistencies can be clearly seen in Table 9.

Table 9: Grad-CAM results from coral images

TABLE IX
GRAD-CAM RESULTS FROM CORAL IMAGES

| Model | Average BBox IoU | mAP50-95 |
| --- | --- | --- |
| YOLOv11_100p | 0.5506 | 0.5064 |
| YOLOv10_100p | 0.4897 | 0.4512 |
| YOLOv9_100p | 0.5033 | 0.313 |
| YOLOv8_100p | 0.4969 | 0.4782 |
| YOLOv11_75p | 0.5078 | 0.4856 |
| YOLOv10_75p | 0.5046 | 0.5501 |
| YOLOv9_75p | 0.5022 | 0.4242 |
| YOLOv8_75p | 0.5135 | 0.5827 |
| YOLOv11_50p | 0.5311 | 0.4752 |
| YOLOv10_50p | 0.4944 | 0.5011 |
| YOLOv9_50p | 0.519 | 0.3158 |
| YOLOv8_50p | 0.5158 | 0.5755 |
| YOLOv11_25p | 0.5309 | 0.4404 |
| YOLOv10_25p | 0.4915 | 0.4974 |
| YOLOv9_25p | 0.5209 | 0.414 |
| YOLOv8_25p | 0.5048 | 0.5372 |

Grad-CAM computes the relevance of each feature map in a convolutional layer by taking the gradient of the output class score with respect to the feature maps. Then it combines and weights the maps to form a heatmap to show where the model placed greater emphasis on the classification. However, YOLO is not a single classification model. Rather, it is a very complex, dense, regression-based object detector that predicts the bounding boxes and confidence score for each class probability. Furthermore, Grad-CAM assumes that the model outputs a single class prediction, but YOLO generates outputs for every individual grid cell. This would often lead to heatmaps highlighting background regions and unrelated parts of the image. A similar condition can be seen in Figure 8.

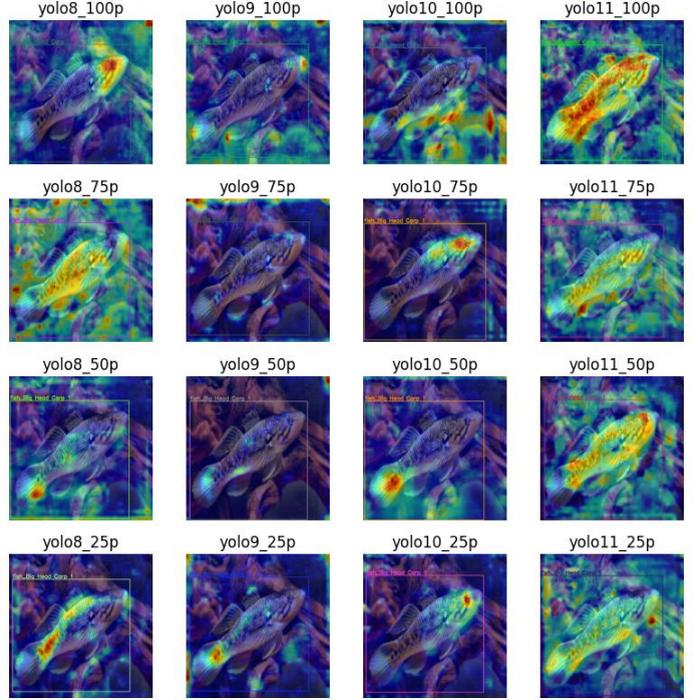

**FIGURE 8.** Grad-CAM from fish images.

From Figure 8, we can see that the models are generally performing better as the models are trained with more data samples. However, it seems like they are pretty unstable, as most models are capturing plenty of noise. Table 10 further presents the Grad-CAM results.

TABLE X
GRAD-CAM RESULTS FROM FISH IMAGES

| Model | Average BBox IoU | mAP50-95 |
| --- | --- | --- |
| YOLOv11_100p | 0.5486 | 0.6256 |
| YOLOv10_100p | 0.5516 | 0.6346 |
| YOLOv9_100p | 0.5852 | 0.5421 |
| YOLOv8_100p | 0.566 | 0.6291 |
| YOLOv11_75p | 0.5739 | 0.6053 |
| YOLOv10_75p | 0.5399 | 0.5964 |
| YOLOv9_75p | 0.5834 | 0.5331 |
| YOLOv8_75p | 0.5628 | 0.6315 |
| YOLOv11_50p | 0.5922 | 0.6188 |
| YOLOv10_50p | 0.5834 | 0.6607 |
| YOLOv9_50p | 0.6049 | 0.5854 |
| YOLOv8_50p | 0.5918 | 0.6651 |
| YOLOv11_25p | 0.6689 | 0.7182 |
| YOLOv10_25p | 0.6271 | 0.6688 |
| YOLOv9_25p | 0.6449 | 0.5969 |
| YOLOv8_25p | 0.6799 | 0.7308 |

The contradiction between the results presented by Grad-CAM and the results presented by the standard evaluation metrics reveals the shortcomings of the Grad-CAM technique in capturing complex models like YOLO.

## V. Conclusion

YOLO has been recognized as a powerful computer vision model capable of both detection and classification. However, most existing benchmarks are based on terrestrial data, raising questions about their performance in underwater environments. This paper evaluates the performance of state-of-the-art YOLO models (YOLOv9–11) to assess their relative effectiveness. Furthermore, Grad-CAM is employed to analyze their visual attention.

The results reveal that while YOLO models can achieve impressive accuracy with limited data samples, the improvements in accuracy between versions are minimal in underwater settings. The most notable enhancement across versions is inference time, with newer models being generally more lightweight. Grad-CAM analysis shows that YOLO models, regardless of version, remain susceptible to noise, often focusing excessively on background elements.

Future work can focus on creating a benchmark for all YOLO models (YOLOv1-11) in underwater settings and analyzing their relative improvements. Furthermore, additional studies need to be conducted on improving explainability metrics for complex multi-task models like YOLO.

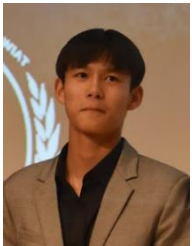
**Gordon Hung** was born in Hsinchu, Taiwan, in 2008. He is a student in Hsinchu County American School and has worked on machine learning and computer vision for over 4 years.

He has interned at the UC Santa Cruz Inter-Networking Research Group, focusing on developing decentralized federated learning frameworks for enhanced temperature prediction. Furthermore, he has been working as an independent researcher to model air pollutants and simulate solutions. Notably, he has presented his work at seven international conferences and published original articles like "Advanced SABLSTM for Comprehensive Scenario Analysis and Simulations of Sweden's Climate Action" (ACM, 2025) and "Accurate Predictive Modeling of Global Coral Reef Bleaching Using Machine Learning" (IEEE, 2025). His current research interests lie in deep learning, environmental engineering, and computer vision.

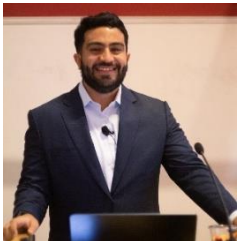
**Ivan Felipe Rodriguez** was born in Bogotá, Colombia, and is a final-year Ph.D. candidate in Cognitive Sciences at Brown University, Providence, RI, USA, where he holds the Presidential Fellowship. He received an M.Sc. in Applied Mathematics from the University of Puerto Rico, Río Piedras, and a B.Sc. in Mathematics from Universidad Sergio Arboleda, Bogotá. His research, conducted under the supervision of Prof. Thomas Serre, focuses on biologically inspired models of vision and decision-making, aiming to align the temporal dynamics and recognition strategies of deep neural networks with human perception and behavior. He was awarded the Presidential Fellowship at Brown University. He has published in top machine learning venues such as NeurIPS, where his work has advanced the development of more interpretable and human-aligned AI systems. His broader research interests include computer vision, computational neuroscience, biologically inspired and interpretable AI, and large-scale applied machine learning.